\documentclass[sigconf]{acmart}
\copyrightyear{2020}
\acmYear{2020}
\setcopyright{acmcopyright}
\acmConference[KDD '20] {26th ACM SIGKDD Conference on Knowledge Discovery and Data Mining}{August 23--27, 2020}{Virtual Event, USA}
\acmBooktitle{26th ACM SIGKDD Conference on Knowledge Discovery and Data Mining (KDD '20), August 23--27, 2020, Virtual Event, USA}
\acmPrice{15.00}
\acmDOI{10.1145/XXXXXX.XXXXXX}
\acmISBN{978-1-4503-7998-4/20/08}

\usepackage{subfigure}
\usepackage{amssymb}

\DeclareMathOperator*{\argmin}{\arg\!\min}
\usepackage{arydshln}

\usepackage{mathtools}
\DeclarePairedDelimiterX{\infdivx}[2]{(}{)}{%
  #1\;\delimsize\|\;#2%
}

\newcommand{\kld}{D_{KL}\infdivx}

\begin{document}

\title{Multi-Source Deep Domain Adaptation with Weak Supervision for Time-Series Sensor Data}

\author{Garrett Wilson}
\orcid{0000-0002-6760-754X}
\email{garrett.wilson@wsu.edu}
\affiliation{%
  \institution{Washington State University}
  \streetaddress{School of Electrical Engineering and Computer Science}
  \city{Pullman}
  \state{WA}
  \postcode{99164}
  \country{USA}}

\author{Janardhan Rao Doppa}
\orcid{0000-0002-3848-5301}
\email{jana.doppa@wsu.edu}
\affiliation{%
  \institution{Washington State University}
  \streetaddress{School of Electrical Engineering and Computer Science}
  \city{Pullman}
  \state{WA}
  \postcode{99164}
  \country{USA}}

\author{Diane J. Cook}
\orcid{0000-0002-4441-7508}
\email{djcook@wsu.edu}
\affiliation{%
  \institution{Washington State University}
  \streetaddress{School of Electrical Engineering and Computer Science}
  \city{Pullman}
  \state{WA}
  \postcode{99164}
  \country{USA}}

\begin{abstract}
Domain adaptation (DA) offers a valuable means to reuse data and models for new problem domains. However, robust techniques have not yet been considered for time series data with varying amounts of data availability. In this paper, we make three main contributions to fill this gap. First, we propose a novel \emph{Convolutional deep Domain Adaptation model for Time Series data (CoDATS)} that significantly improves accuracy and training time over state-of-the-art DA strategies on real-world sensor data benchmarks. By utilizing data from multiple source domains, we increase the usefulness of CoDATS to further improve accuracy over prior single-source methods, particularly on complex time series datasets that have high variability between domains. Second, we propose a novel \emph{Domain Adaptation with Weak Supervision (DA-WS)} method by utilizing weak supervision in the form of target-domain label distributions, which may be easier to collect than additional data labels. Third, we perform comprehensive experiments on diverse real-world datasets to evaluate the effectiveness of our domain adaptation and weak supervision methods. Results show that CoDATS for single-source DA significantly improves over the state-of-the-art methods, and we achieve additional improvements in accuracy using data from multiple source domains and weakly supervised signals.
\end{abstract}

\ccsdesc[500]{Computing methodologies~Transfer learning}
\ccsdesc[300]{Computing methodologies~Unsupervised learning}
\ccsdesc[300]{Computing methodologies~Adversarial learning}
\ccsdesc[300]{Computing methodologies~Neural networks}
\ccsdesc[300]{Mathematics of computing~Time series analysis}

\keywords{transfer learning, domain adaptation, time series, human activity recognition, weak supervision}

\maketitle

\section{Introduction}

Time series sensor data abound in many real-world settings including human activity recognition \cite{anguita2013public}, sleep stage classification \cite{zhao2017icml}, gesture recognition \cite{uWaveDataset}, speech recognition \cite{hosseiniasl2019augmented}, and diagnosis and mortality prediction from medical data \cite{purushotham2017variational}. Labeling data in these situations is expensive and sometimes infeasible. One way to reduce labeling effort is to design unsupervised domain adaptation techniques that leverage the labeled data from one or more source domains and unlabeled data from a new target domain to build a classifier for the target domain \cite{ganin2016jmlr,xie2017nips}.

While unsupervised domain adaptation methods have been designed for image data, very limited work has focused on adaptation approaches for time series data \cite{wilson2019survey}. A few time series methods have been introduced. However, these prior approaches utilize recurrent neural networks (RNNs) that can be very slow to train for reasonable-sized time series arising in real-world problems. While researchers have found that convolutional neural networks (CNNs) can achieve the same accuracy as RNNs while being trained and evaluated much faster \cite{miller2018stable,bai2018empirical}, previously-proposed domain adaptation network architectures are incompatible with time series data.

\begin{figure*}
    \centering
    \includegraphics[width=0.99\linewidth]{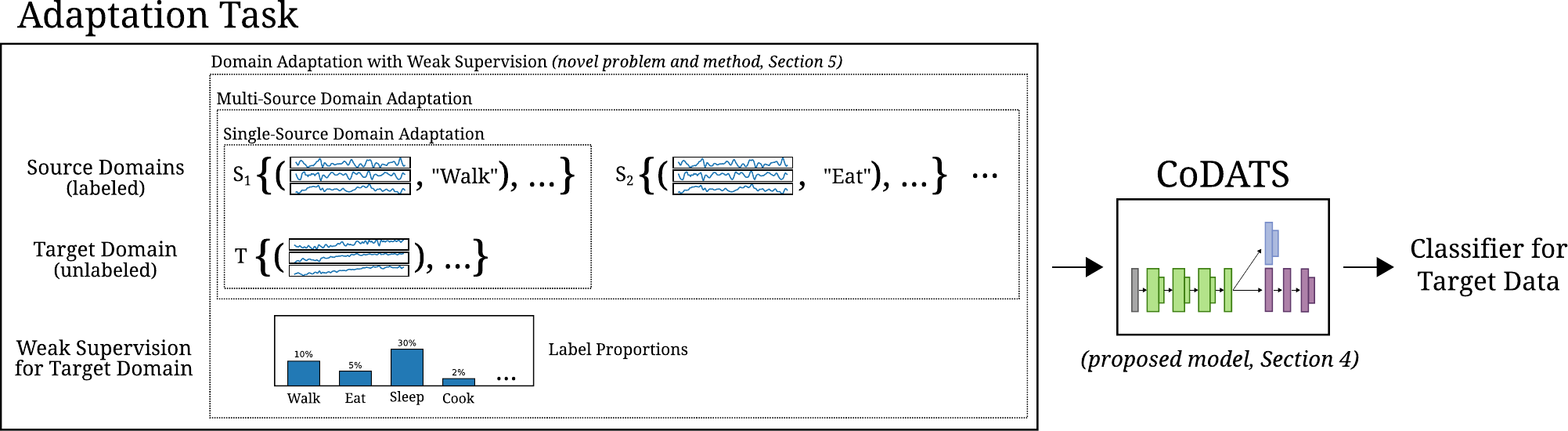}
    \Description{Data for the adaptation task (single-source, multi-source, or multi-source with weak supervision) is input to our proposed CoDATS model, which outputs a classifier for the target data.}
    \caption{Illustration of our proposed CoDATS model that supports single- and multi-source domain adaptation in addition to our novel problem setting of domain adaptation with weak supervision.}
    \label{fig:overview}
\end{figure*}

In this paper, we propose a new model: \textbf{Co}nvolutional deep \textbf{D}omain \textbf{A}daptation model for \textbf{T}ime \textbf{S}eries data (CoDATS). CoDATS couples technical principles from domain-invariant domain adaptation with a network design that is more efficient, accurate, time-series compatible, and extensible than prior work. The CoDATS architecture exhibits three important features. First, it leverages existing domain-invariant domain adaptation methods to operate on time series data. Second, it outperforms existing single-source time series adaptation models. Third, it is readily extensible to additional situations including when data from multiple source domains is available, which is particularly helpful for complex time series datasets having high variability between domains, and when the target-domain label distribution is available, which may be easier to collect than additional time series data labels.

While utilizing unlabeled target-domain data in unsupervised domain adaptation is one way of reducing labeling effort, another way is to use weakly-supervised information that is relatively easy to acquire. Obtaining labels for time series sensor data is more challenging than for image data. For example, cat vs. dog image classification labels can be easily obtained after data collection by having a person look at each image and determine if the image is of a cat or a dog. In contrast, for time series human activity recognition, it is much more difficult to identify what activity a human was performing by looking at raw accelerometer, gyroscope, and magnetometer sensor data. Thus, labels for time series sensor data are instead typically recorded while performing the activity \cite{minor2017tkde,froehlich2007myexperience}, which greatly limits the number of gathered labels because of the additional burden from interrupting a person's activities. However, other possibilities exist for obtaining information in the form of weak supervision. For activity recognition, it may be easy for each participant to self-report what proportion of the time they performed each activity. For example, providing an estimate of how many hours a day they spend cooking is easier than labeling each data instance of cooking. We formulate this new problem setting of \textbf{D}omain \textbf{A}daptation with \textbf{W}eak \textbf{S}upervision (DA-WS) and develop a novel method to effectively utilize weak supervision in the form of label proportions. The key idea is to constrain the space of model parameters to those which approximately matches the label proportions on unlabeled data from the target domain.

To validate our proposed CoDATS model and weak-supervision method, we performed comprehensive experiments on diverse real-world time series benchmarks including gesture recognition and human activity recognition. We compare CoDATS with prior single-source time series methods and observe that CoDATS dramatically outperforms previous approaches to time series domain adaptation. Additionally, we demonstrate how CoDATS can further improve accuracy by utilizing data from multiple sources. We also find that coupled with our proposed CoDATS model, our DA-WS method yields additional improvements in accuracy.

\vspace{0.5ex}

\noindent {\bf Contributions.} We make three key contributions, as summarized in Figure~\ref{fig:overview}. {\bf 1)} We develop a new time-series compatible model referred as CoDATS to improve both accuracy and computational-efficiency when compared to prior work on single-source DA. CoDATS supports utilizing data from multiple sources to further improve accuracy. {\bf 2)} We formulate a new weak supervision problem to leverage target-domain label proportions when available and propose a novel method referred as DA-WS to effectively solve it. {\bf 3)} We perform comprehensive experimental evaluation on multiple challenging real-world benchmarks to show the efficacy of our CoDATS model and weak-supervision method over state-of-the-art.\footnote{Code is available at: \url{https://github.com/floft/codats}}

\section{Related Work}

A large body of prior work has contributed numerous strategies for single-source domain adaptation of image data. These methods learn a domain-invariant feature representation \cite{kang2019contrastive,ganin2016jmlr,sankaranarayanan2018cvpr} or perform domain mapping \cite{hoffman2018icml,bousmalis2017cvpr,hong2018cvpr}. Some researchers explored using separate normalization statistics for each domain \cite{li2018}, building ensembles \cite{french2018iclr}, or pushing decision boundaries into lower-density regions \cite{shu2018vada,kumar2018nips}. While multi-source domain adaptation, in which multiple source domains are adapted to a single target, has not received the same level of attention, a few methods have been proposed. One straightforward method simply combines multiple domains with labeled data together as a single ``source'' domain \cite{sun2015multisourcesurvey}. More sophisticated methods take advantage of differences among source domains or the relationship between sources and target. Zhao et al. \cite{zhao2018multisource} include a separate domain classifier for each source domain and compute a loss based on the lowest domain error among these classifiers. Xie et al. \cite{xie2017nips} instead propose using a multi-class output for a single domain classifier, which scales better as the number of source domains increases. These previous single-source and multi-source adaptation algorithms cannot be directly applied to time series data without a time-series compatible model architecture, such as our proposed CoDATS model.

While less numerous, researchers have introduced a few time series adaptation methods. Specifically, variational recurrent adversarial deep domain adaptation (VRADA) and recurrent domain adversarial neural network (R-DANN) explore RNNs as feature extractors. These RNNs can be combined with an adversary to make a time series representation domain invariant \cite{purushotham2017variational}. R-DANN employs a long short-term memory (LSTM) network, and VRADA uses a variational RNN. These were tested on time series medical data, while a variation employing gated recurrent units (GRUs) predicted driving maneuvers \cite{tonutti2019robust}.

A few domain adaptation methods have tackled specialized time series tasks. These include instance weighting for anomaly detection \cite{vercruyssen2017transfer} and a domain-invariant feature learner for inertial tracking that combined an autoencoder, a generative adversarial network (GAN), and RNNs \cite{chen2019motiontransformer}. Another domain-invariant method combined CNNs and RNNs to perform domain generalization for classification of sleep stages from radio-frequency data \cite{zhao2017icml}. In a related method, source datasets with differing label spaces were transferred to one target dataset using pre-training and fine-tuning \cite{fawaz2018tltimeseries}.

We include the two most related approaches, VRADA and R-DANN, as baseline models in our empirical study validating our proposed CoDATS model since they were developed for time series classification and thus can be directly used for comparison. However, we do not include the others since they solve related but different problems or are incompatible with time series data.

Our proposed domain adaptation with weak supervision approach is inspired in part by the posterior regularization idea from Ganchev et al., in which learning is guided to meet certain constraints on the unlabeled data \cite{ganchev2010posterior}. However, our work differs from that of Ganchev et al. in three key ways. First, they focus on developing methods for natural language processing, whereas we focus on developing appropriate methods and models compatible with time series. Second, they develop weak supervision for use alone or in tandem with some additional labeled data for semi-supervised learning, whereas we focus on coupling weak supervision with domain adaptation. Third, they develop a weak supervision method based on expectation maximization, whereas because we are using deep neural networks, we develop a weak supervision objective that instead takes the form of a differentiable regularization term added to the loss function. A related regularizer was used by Jiang et al. \cite{jiang2018towards}, but they estimate label proportions on the source domain data rather than using target-domain weak supervision.

A few other weak supervision methods have been proposed in different contexts. Hu et al. \cite{hu2017learning} learn human activity recognition models from video data that may have multiple, uncertain, or incomplete labels. We instead propose a form of weak supervision better suited to streams of raw sensor data that cannot be easily visually identified. Huang et al. \cite{huang2012biased} incorporate expert-provided biases into a hidden Markov model (HMM) for domain adaptation in a natural language processing context. The biases provide information about a specific task in an application domain (akin to feature engineering) rather than information about the target domain data, which may differ for each target domain. In the context of semantic segmentation of images, a variety of work has explored using weak supervision \cite{pathak2015iccv,hong2016cvpr}. However, such weak supervision takes the form of image labels or bounding boxes, which cannot be directly applied to time series sensor data. Despite the fundamental differences between these methods and the problem we are addressing, the performance gains suggest that weak supervision in a time series domain adaptation context may similarly yield an improvement in accuracy.

\section{Problem Setup}

Domain adaptation reuses labeled data from one domain to create a classifier for a different but related domain. In time series data settings such as human activity recognition, this is a common problem. For example, activity-labeled sensor data may exist for person 1 while only unlabeled data is available for person 2. Because of inherent differences (e.g., different activity patterns, sensor positions, or sampling rates), a classifier trained on person 1's data will likely not perform well for person 2. However, we can create a feature extractor that produces domain-invariant features. A classifier trained on these features will generalize better to person 2 because the features are similar in both domains.

Additionally, we may have other sources of information about person 2's activities. For example, they may self-report how much time they spend on each activity, which can be interpreted as label proportions for the target domain. This can be used as a constraint or regularizer for the learned model: learn a model that is domain invariant but also consistent with the known label proportions.

Below we provide definitions for two cases of unsupervised domain adaptation: single-source domain adaptation and multi-source domain adaptation. Following this, we formulate our novel problem setting of domain adaptation with weak supervision.

\vspace{0.5ex}

\noindent {\bf Single/Multi-Source Domain Adaptation.}
Formally, given input data $X$ with $L$ labels $Y = \{ 1, 2, \dots, L \}$ and $n$ source domains, then we have several distributions over the input and label space $X \times Y$: $n$ source domain distributions $\mathcal{D}_{S_i}$ for $i \in \{ 1, 2, \dots, n \}$ and a target domain distribution $\mathcal{D}_T$. During training, we draw labeled samples i.i.d. from each source domain distribution $\mathcal{D}_{S_i}$ and unlabeled samples i.i.d. from $\mathcal{D}_T^X$ (the marginal distribution of $\mathcal{D}_T$ over $X$). This gives us $s_i$ source samples $S_i$ for each source domain $i \in \{ 1, 2, \dots, n \}$ and $t$ target samples $T$:
\begin{equation}\label{eq:msda1}
    S_i = \{ (\textbf{x}_j, y_j) \}^{s_i}_{j=1} \sim \mathcal{D}_{S_i}, \quad \forall i \in \{ 1, 2, \dots, n \}
\end{equation}
\begin{equation}\label{eq:msda2}
    T = \{ (\textbf{x}_j) \}^t_{j=1} \sim \mathcal{D}_T^X
\end{equation}

Single-source domain adaptation corresponds to the case in which $n=1$ and multi-source domain adaptation corresponds to the case where $n > 1$. During evaluation, we test the learned model on $t$ samples $T_{test}$ drawn i.i.d. from $\mathcal{D}_T$ and not seen during training (we need ground-truth labels to compute accuracy).
\begin{equation}\label{eq:msda3}
  T_{test} = \{ (\textbf{x}_j, y_j) \}^t_{j=1} \sim \mathcal{D}_T
\end{equation}

In this paper, we specifically investigate the case where $X$ represents time series data. Time series can be either univariate or multivariate. If univariate, then $X = [x_1, x_2, \dots, x_H]$ consists of $H$ ordered real values. If multivariate, then $X = [X^1, X^2, \dots, X^K]$ consists of $K$ univariate time series, each of which contains $H$ ordered real feature values \cite{fawaz2019review}. For example, these features may be accelerometer x, y, and z values, each of which is a time series of real values.

\vspace{0.5ex}

\noindent {\bf Domain Adaptation with Weak Supervision.} For weak supervision with known target-domain label proportions, the setup adheres to the above domain adaptation definition, except additional information is now available for training. Formally, we are provided with a discrete probability distribution $P(Y=y)$ giving the probability $p_y$ that a target-domain example will have the label $y$:
\begin{equation}\label{eq:labelproportions}
    Y_{true}(y) = P(Y=y) = p_y, \quad \forall y \in \{ 1, 2, \dots, L \}
\end{equation}

Note that because this is a probability distribution, $\sum_{y=1}^L p_y = 1$ and $p_y \geq 0$ for all $y \in \{ 1, 2, \dots, L \}$.

\section{CoDATS for Domain Adaptation}

We propose a time series domain adaptation model to improve adaptation performance. Here, we first summarize the proposed adaptation method that supports both single-source and multi-source domain adaptation. Next, we detail the network architecture we develop that is compatible with time series data.

\begin{figure}
    \subfigure[(a)][Training]{\begin{minipage}{1.0\linewidth}
        \centering
        \includegraphics[width=0.9\linewidth]{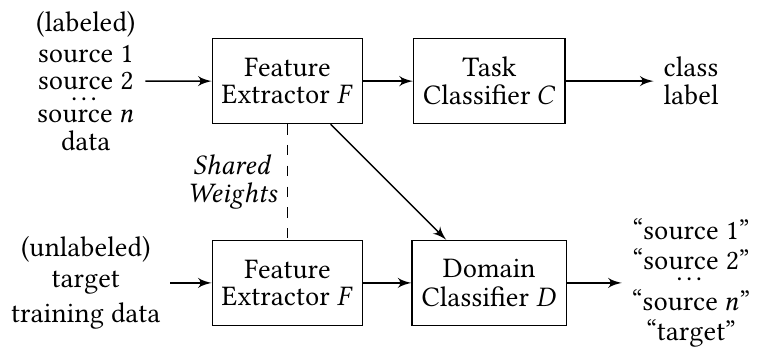}
    \end{minipage}}
    \subfigure[(b)][Testing]{\begin{minipage}{1.0\linewidth}
        \centering
        \includegraphics[width=0.9\linewidth]{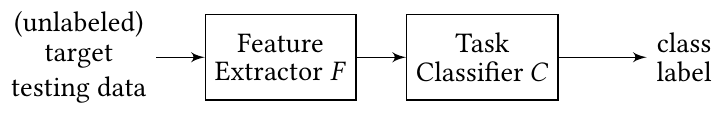}
    \end{minipage}}
    \Description{During training, labeled data from each source domain is fed through the feature extractor then the task classifier. Unlabeled target data is fed through the feature extractor. The representations are aligned with a domain classifier adversary. During testing, target data is fed through the feature extractor and the task classifier.}
    \caption{CoDATS training and testing setup. Its goal is to learn a domain-invariant feature representation during training that is then used during testing.}
    \label{fig:codats_method}
\end{figure}

\subsection{Adaptation Method}

We propose a domain-invariant training method for CoDATS. The goal of domain-invariant methods is to learn a feature extractor network that produces domain-invariant features. Such features can be learned with a domain classifier (or ``discriminator'') acting as an adversary. If a classifier trained on these domain-invariant features performs well in the source domain, then this classifier may generalize well to the target domain due to the domain-aligned feature distributions. Ultimately, generalization also depends on the level of similarity between domains. If the domains are too distinct, learning domain-invariant features can increase error \cite{ben2010ml}. The most common domain-invariant method is the domain adversarial neural network (DANN) \cite{ganin2016jmlr}, which is the basis of the prior time-series adaptation works, VRADA and R-DANN.  However, note that DANN only works for single-source domain adaptation.

Xie et al. \cite{xie2017nips} extend DANN to a more general framework, supporting invariance from factors or traits other than binary domain labels. While their framework allows for multinomial variables, continuous variables, or variables with structure such as parse trees, for multi-source domain adaptation (or single-source, which is a special case) we are particularly interested in the case of multinomial variables. Thus, we base our adaptation method on this framework.

Our training setup illustrated in Figure~\ref{fig:codats_method} consists of three neural networks: a feature extractor $F$, a task classifier $C$, and a domain classifier $D$. The feature extractor is updated in an adversarial manner, with two competing objectives. The feature extractor and task classifier are updated such that the task classifier correctly classifies labeled source data. Similarly, the domain classifier is updated in such a way that it accurately classifies data as coming from the correct domain. At the same time, the feature extractor is also updated such that the domain classifier cannot classify which domain the data originated from (the adversarial component). The adversarial step is performed via a gradient reversal layer (GRL) \cite{ganin2016jmlr} placed in the network between the feature extractor and domain classifier. The GRL flips the gradient during back propagation when updating the weights to achieve this effect.

Formally, given a discriminator $D(\cdot;\theta_d)$ with parameters $\theta_d$, a feature extractor $F(\cdot;\theta_f)$ with parameters $\theta_f$, a task classifier $C(\cdot;\theta_c)$ with parameters $\theta_c$, labeled source domain data $\mathcal{D}_{S_i}$ from a source domain $i \in \{ 1, 2, \dots n \} $, unlabeled target domain data $\mathcal{D}_T^X$, two multi-class cross entropy losses $\mathcal{L}_y$ and $\mathcal{L}_d$ for the labels and domains respectively, and domain labels for each example where the target domain is labeled $d_T = 0$ and the source domains are labeled $d_{S_i} = i$ for $i \in \{ 1, 2, \dots n \}$, then the two competing objectives are:
\begin{align}\label{eq:codats1}
\argmin_{\theta_f,\theta_c} \quad
    \sum_{i=1}^{n} \mathbb{E}_{(\mathbf{x},y) \sim \mathcal{D}_{S_i}} \big[ & \mathcal{L}_y(C(F(\mathbf{x})), y) - \mathcal{L}_d(D(F(\mathbf{x})), d_{S_i}) \big] \nonumber \\
    - \mathbb{E}_{\mathbf{x} \sim \mathcal{D}_T^X} \big[ & \mathcal{L}_d(D(F(\mathbf{x})), d_T) \big]
\end{align}
\begin{align}\label{eq:codats2}
\argmin_{\theta_d} \quad
    \sum_{i=1}^{n} \mathbb{E}_{(\mathbf{x},y) \sim \mathcal{D}_{S_i}} \big[ & \mathcal{L}_d(D(F(\mathbf{x})), d_{S_i}) \big] \nonumber \\
+ \mathbb{E}_{\mathbf{x} \sim \mathcal{D}_T^X} \big[ & \mathcal{L}_d(D(F(\mathbf{x})), d_T) \big]
\end{align}

To summarize, Equation~\ref{eq:codats1} updates the parameters of $F$ and $C$ such that $C$ predicts the correct source task labels and $D$ predicts the incorrect domain labels (the $\mathcal{L}_d$ terms are negated), while Equation~\ref{eq:codats2} updates $D$ to correctly predict the domain labels.

However, using the gradient reversal layer, Equations~\ref{eq:codats1} and \ref{eq:codats2} can be combined together into one step. The gradient reversal layer can be represented as $\mathcal{R}(\mathbf{x})$ with different forward and backward propagation behavior, where $\mathbf{I}$ is the identity matrix and $\lambda$ is a constant (possibly with a specified schedule during training):
\begin{equation}\label{eq:grl}
\mathcal{R}(\mathbf{x}) = \mathbf{x}; \quad \frac{d\mathcal{R}}{d\mathbf{x}} = -\lambda\mathbf{I}
\end{equation}

Then, the optimization step becomes:
\begin{align}\label{eq:grlFormulation}
\argmin_{\theta_f,\theta_c,\theta_d} \quad
    & \sum_{i=1}^{n} \mathbb{E}_{(\mathbf{x},y) \sim \mathcal{D}_{S_i}} \big[ \mathcal{L}_y(C(F(\mathbf{x})), y)
        + \mathcal{L}_d(D(\mathcal{R}(F(\mathbf{x}))), d_{S_i}) \big] \nonumber \\
    & + \mathbb{E}_{\mathbf{x} \sim \mathcal{D}_T^X} \big[ \mathcal{L}_d(D(\mathcal{R}(F(\mathbf{x}))), d_T) \big]
\end{align}

Equation~\ref{eq:grlFormulation} yields an adaptation method capable of handling both single-source ($n=1$) and multi-source ($n>1$) domain adaptation.

\subsection{Model}
We now need to design time-series compatible network architectures capable of handling the shape and temporal nature of time series data. In contrast to image data which are typically represented by three dimensions (height, width, number of channels) having spatial relationships along two dimensions, time series data typically have two dimensions (time series length, number of features) with temporal relationships between sensor data across time. The developed network must be capable of learning dependencies along this time dimension.

Prior works used RNNs for time series because of their ability to handle sequential data. However, RNNs pose a number of challenges: long-term dependencies are inhibited because gradients backpropagated through time tend to vanish and training can fail if gradients explode. Gating mechanisms have been developed to partially address vanishing gradients \cite{bai2018empirical}, and exploding gradients can be partially resolved by gradient clipping \cite{pascanu2013difficulty}. Others have tried developing superior RNNs, but these have been found to not outperform LSTMs for many tasks \cite{jozefowicz2015empirical,greff2017lstm,melis2017state}. An alternative direction for improving sequential models is instead employing feed-forward networks like CNNs, using 1D convolutions along the time axis. Stable RNNs have provably-good feed-forward approximations \cite{miller2018stable}, and researchers have empirically demonstrated the benefit of CNNs on sequential data, for example, performing better at long-term memory than RNNs \cite{bai2018empirical}.

\begin{figure}
    \centering
    \includegraphics[width=0.78\linewidth]{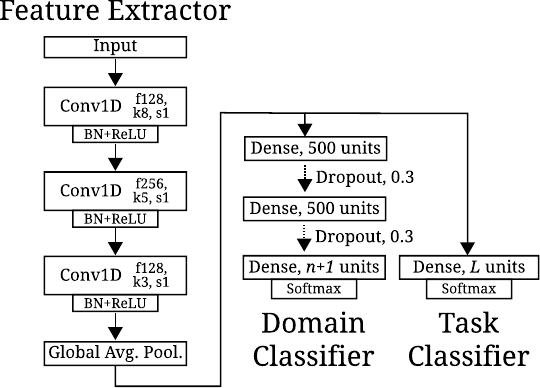}
    \Description{The feature extractor consists of three 1D convolutional layers of filter sizes 128, 256, and 128 and kernel sizes 8, 5, and 3 each followed by batch norm and ReLU. Then, a global average pooling layer. The domain classifier consists of three dense layers of 500 units with 0.3 dropout. Then, a softmax. The task classifier consists of a single dense layer with as many units as class labels. Then, a softmax.}
    \caption{CoDATS model architecture. Each convolution layer uses same padding. f$x$ indicates $x$ filters, k$x$ a kernel of size $x$, and s$x$ a stride of $x$. There are $n+1$ domain classifier outputs due to $n$ source domains and 1 target domain.}
    \label{fig:codats_model}
\end{figure}

While time-series adaptation traditionally employs RNNs, we hypothesize that both accuracy and computation cost can benefit from a CNN architecture. We propose the CoDATS networks summarized in Figure~\ref{fig:codats_model}. The feature extractor consists of a fully convolutional network (FCN) \cite{wang2017strongbaseline} that is comparable to more computationally-expensive, state-of-the-art ensemble time series classification models \cite{wang2017strongbaseline,fawaz2019review}. A similar FCN model has previously been used for other time series transfer learning tasks \cite{fawaz2018tltimeseries}. The single dense layer (the last layer) of the FCN then acts as the task classifier. The domain classifier acting as the adversary during training consists of a multi-layer perceptron (MLP) model \cite{wang2017strongbaseline}. We employ MLP rather than additional 1D convolutions since following the global average pooling layer in the feature extractor, there is no longer any time dimension. Thus, additional 1D convolutions along the time axis are not possible. Other works have similarly used multiple dense layers for a domain classifier \cite{ganin2016jmlr,purushotham2017variational}.

Similar to an RNN, the proposed network can handle variable-length time series data. RNNs handle variable-length data by processing each time step sequentially, whereas the proposed CNN network handles variable-length data with the global average pooling layer. While the data in our experiments are of a fixed length, the proposed model is not limited to such situations.

\section{DA with Weak Supervision}

We propose a novel method for domain adaptation with weak supervision (DA-WS) based on our CoDATS model. The general principle behind our learning approach is to use the weak supervision information as constraints to efficiently search for good model parameters. We can devise specific learning methods depending on the form of weak supervision by appropriately instantiating this general principle. DA-WS incorporates the known target-domain label proportion information during training by searching for model parameters whose predicted labels on unlabeled target domain data approximately matches with the given label proportions. Mathematically, this is done by introducing a differentiable regularization term to the training objective.

From our formalization of the problem in Equation~\ref{eq:labelproportions}, we have the true label proportions $Y_{true}(y)$ available for the target domain. Now let us define the predicted target-domain distribution from the CoDATS task classifier. Let $C(F(\mathbf{x}))_y$ denote the element of the task classifier's softmax prediction corresponding to label $y$. Then, the predicted label distribution can be represented as a probability distribution $P(Y=y|X=\mathbf{x})$, indicating the probability that a target-domain instance $\mathbf{x} \sim \mathcal{D}_T^X$ will have label $y$:
\begin{equation}\label{eq:pred_labelproportions}
    Y_{pred}(y) = P(Y_{pred}=y) = C(F(\mathbf{x}))_y, \quad \forall y \in \{ 1, 2, \dots, L \}
\end{equation}

We do \textit{not} want each individual prediction to have output probabilities matching $Y_{true}(y)$, which would regularize to always predict the majority class (the maximum probability in the $Y_{true}(y)$ discrete distribution). Rather, we want the expected value computed over a large number of predictions (e.g., a mini-batch) to follow this distribution. We want $C$ to predict a particular class for each $\mathbf{x}$, but on average the predictions should follow $Y_{true}(y)$. This gives us the estimated distribution $\tilde{Y}_{pred}(y)$ over multiple predictions:
\begin{equation}\label{eq:est_pred_distribution}
    \tilde{Y}_{pred}(y) = \mathbb{E}_{\mathbf{x} \sim \mathcal{D}_T^X} \big[ Y_{pred}(y) \big] = \mathbb{E}_{\mathbf{x} \sim \mathcal{D}_T^X} \big[ C(F(\mathbf{x})) \big]
\end{equation}

During training, we can align the predicted distribution $\tilde{Y}_{pred}(y)$ with the known distribution $Y_{true}(y)$ with a distance measure such as Kullback-Leibler (KL) divergence. Thus, we propose adding the following weak supervision regularization term to the loss:
\begin{align}\label{eq:reg}
    \mathcal{L}_{WS}
    & = \kld*{Y_{true}(y)}{\tilde{Y}_{pred}(y)} \nonumber \\
    & = \kld*{Y_{true}}{\mathbb{E}_{\mathbf{x} \sim \mathcal{D}_T^X} \big[ C(F(\mathbf{x})) \big]}
\end{align}

Combining our regularization term in Equation~\ref{eq:reg} with the CoDATS training objective in Equation~\ref{eq:grlFormulation}, the entire training objective for DA-WS becomes:
\begin{align}\label{eq:dawsGrlFormulation}
\argmin_{\theta_f,\theta_c,\theta_d} \quad
    & \sum_{i=1}^{n} \mathbb{E}_{(\mathbf{x},y) \sim \mathcal{D}_{S_i}} \big[ \mathcal{L}_y(C(F(\mathbf{x})), y)
        + \mathcal{L}_d(D(\mathcal{R}(F(\mathbf{x}))), d_{S_i}) \big] \nonumber \\
    & + \mathbb{E}_{\mathbf{x} \sim \mathcal{D}_T^X} \big[ \mathcal{L}_d(D(\mathcal{R}(F(\mathbf{x}))), d_T) \big] \nonumber \\
    & + \kld*{Y_{true}}{\mathbb{E}_{\mathbf{x} \sim \mathcal{D}_T^X} \big[ C(F(\mathbf{x})) \big]}
\end{align}

Altogether, this objective finds model parameters that: 1) predict correct task labels using the labeled data from the source domains, 2) produce a domain-invariant feature representation among all domains using the unlabeled target-domain data, and 3) predict target-domain task labels that on average align well with the known target-domain label proportions.

Note that this regularizer only has an effect when the label distribution differs between the source and target domains. If the distributions are the same, then when training with empirical risk minimization, $C(F(\mathbf{x}))$ will already produce a predicted label distribution aligning with that of the source(s), making the KL divergence term zero. We only gain information from knowing the target-domain label distribution $Y_{true}$ if it differs from the sources' label distribution, and thus only in such situations will we see DA-WS yield an improvement. Indeed, our experimental results corroborate this proposition.

\section{Experimental Setup}

First, to evaluate our proposed model, we compare CoDATS with relevant prior methods and several baselines on diverse time series adaptation benchmarks. We hypothesize that CoDATS will yield an improvement in accuracy while also reducing training time. We choose accuracy as the evaluation metric to afford comparison with prior adaptation work. While we initially compare on single-source adaptation because this is the problem studied by prior work, we also demonstrate that our model supports multi-source adaptation, which we hypothesize can yield additional performance gains.

Second, we perform experiments utilizing our proposed domain adaptation with weak supervision (DA-WS) method. We hypothesize that DA-WS coupled with our CoDATS model will provide further improvements in accuracy.

\subsection{Datasets}

A requirement for selecting datasets to test domain adaptation is that the dataset includes a property for splitting the data into multiple domains. Such properties could include different data collection days or different generating processes (e.g., different persons in the case of human activity recognition data). The multi-variate time series datasets we select include a participant identifier, and we use this feature to split data into multiple domains. Thus, our adaptation problems consist of the realistic use-case adapting a model from one or more participants' data to another participant's data.

We include the Human Activity Recognition (HAR) dataset \cite{anguita2013public} due to its popularity in time series research. It contains accelerometer, gyroscope, and estimated body acceleration data from 30 participants. We also include the Heterogeneity Human Activity Recognition Dataset (HHAR) dataset \cite{stisen2015smartdevices}, which is more diverse than HAR, including accelerometer data from 31 smartphones of different manufacturers, models, etc. positioned in various orientations. Additionally, we include 33 participants' accelerometer data from the WISDM activity recognition (WISDM AR) dataset \cite{kwapisz2011wisdmar}. Finally, we include a gesture recognition dataset (uWave) containing accelerometer data from 8 participants performing various hand gestures \cite{uWaveDataset}. In the single-source experiments, we pick 10 random pairs of participants for each dataset. In the multi-source experiments, for each value of $n$, we pick 3 random subsets of source domains for each of the 10 random target domains. See the Appendix for more details.

\subsection{CoDATS Model}

\noindent {\bf Single-Source Domain Adaptation.} In our evaluation, we compare CoDATS with several single-source domain adaptation baselines. First, we include no adaptation as an approximate lower bound (\texttt{No Adaptation}). The lower bound gives a rough estimate of the domain gap between the source and target distributions and indicates how hard it will be to perform domain adaptation. We expect that domain adaptation methods will exceed this lower bound, depending on the extent of the domain gap. Similarly, we include models trained directly on labeled target data as an approximate upper bound (\texttt{Trained on Target}). Assuming there is enough target data, then we expect these upper-bound models to perform well when trained on this labeled data. As explained by the theory of Ben-David et al. \cite{ben2010ml}, given enough target labels, not using any of the source domain data is actually preferable because of the possibility of negative transfer. Finally, we include \texttt{R-DANN} and \texttt{VRADA} \cite{purushotham2017variational} as single-source time series adaptation baselines using their respective network architectures. We compare these baselines with our CoDATS model first for single-source domain adaptation to demonstrate the superiority of our proposed model for time series.

\vspace{0.5ex}

\noindent {\bf Multi-Source Domain Adaptation.} As with single-source domain adaptation, we include the baselines \texttt{No Adaptation} and \texttt{Trained on Target} in the multi-source domain adaptation experiments. This set of experiments demonstrate the additional performance gains achievable by utilizing data from multiple people. Our proposed CoDATS model supports such multi-source domain adaptation, whereas prior methods do not.

\subsection{Domain Adaptation with Weak Supervision}

For the domain adaptation with weak supervision problem setting, we are given additional target-domain label proportion information. We simulate being given this information by estimating the label proportions of the target domain data from only its training set. We evaluate our \texttt{DA-WS} method with both a single source domain and also multiple source domains. Since this is a novel problem setting, we do not compare with prior works but rather compare with \texttt{No Adaptation}, \texttt{Trained on Target}, and \texttt{CoDATS}. By comparing with these domain adaptation methods that do not make use of the additional weak supervision information, we determine whether we can successfully utilize this additional information to improve model performance. Note that because the DA-WS method also uses the CoDATS model, we use the term \texttt{CoDATS} to refer to the experiments using the model without weak supervision and the term \texttt{CoDATS-WS} to refer to the experiments using the model with weak supervision.

\begin{table*}
\centering
\begin{small}
\begin{tabular}{cccccccc}
\toprule
Problem & No Adaptation & R-DANN & VRADA & \textit{CoDATS} & \textit{CoDATS-WS} & Train on Target \\
\midrule
HAR 2 $\rightarrow$ 11 & \textbf{83.3 $\pm$ 0.7} & 80.7 $\pm$ 5.2 & 64.1 $\pm$ 5.6 & 74.5 $\pm$ 4.5 & 74.5 $\pm$ 6.0 & 100.0 $\pm$ 0.0 \\
HAR 7 $\rightarrow$ 13 & 89.9 $\pm$ 3.6 & 75.3 $\pm$ 5.8 & 78.3 $\pm$ 5.2 & \textbf{\underline{96.5 $\pm$ 0.7}} & \textbf{\underline{96.5 $\pm$ 0.7}} & 100.0 $\pm$ 0.0 \\
HAR 12 $\rightarrow$ 16 & 41.9 $\pm$ 0.0 & 35.1 $\pm$ 2.9 & 61.7 $\pm$ 7.5 & \textbf{77.5 $\pm$ 0.6} & 75.2 $\pm$ 3.5 & 100.0 $\pm$ 0.0 \\
HAR 12 $\rightarrow$ 18 & 90.0 $\pm$ 1.7 & 74.9 $\pm$ 0.6 & 74.4 $\pm$ 6.7 & \textbf{\underline{100.0 $\pm$ 0.0}} & \textbf{\underline{100.0 $\pm$ 0.0}} & 100.0 $\pm$ 0.0 \\
HAR 9 $\rightarrow$ 18 & 31.1 $\pm$ 1.7 & 56.6 $\pm$ 6.4 & 59.8 $\pm$ 10.1 & \textbf{\underline{85.8 $\pm$ 1.7}} & 76.7 $\pm$ 6.8 & 100.0 $\pm$ 0.0 \\
HAR 14 $\rightarrow$ 19 & 62.0 $\pm$ 4.3 & 71.3 $\pm$ 2.4 & 64.4 $\pm$ 4.7 & 72.2 $\pm$ 27.2 & \textbf{\underline{98.6 $\pm$ 1.1}} & 100.0 $\pm$ 0.0 \\
HAR 18 $\rightarrow$ 23 & \textbf{89.3 $\pm$ 5.0} & 78.2 $\pm$ 6.4 & 72.9 $\pm$ 6.0 & 86.2 $\pm$ 0.6 & \textbf{89.3 $\pm$ 1.1} & 100.0 $\pm$ 0.0 \\
HAR 6 $\rightarrow$ 23 & 52.9 $\pm$ 2.3 & 79.1 $\pm$ 2.7 & 78.2 $\pm$ 6.4 & \textbf{\underline{94.7 $\pm$ 1.1}} & 94.2 $\pm$ 1.3 & 100.0 $\pm$ 0.0 \\
HAR 7 $\rightarrow$ 24 & 94.4 $\pm$ 2.7 & 84.8 $\pm$ 6.9 & 93.9 $\pm$ 0.6 & \textbf{100.0 $\pm$ 0.0} & 99.1 $\pm$ 0.6 & 100.0 $\pm$ 0.0 \\
HAR 17 $\rightarrow$ 25 & 57.3 $\pm$ 5.5 & 66.3 $\pm$ 5.8 & 52.0 $\pm$ 1.1 & \underline{96.7 $\pm$ 1.5} & \textbf{\underline{97.6 $\pm$ 1.0}} & 100.0 $\pm$ 0.0 \\
\hdashline
HAR Average & 69.2 $\pm$ 21.8 & 70.2 $\pm$ 14.0 & 70.0 $\pm$ 11.4 & \underline{88.4 $\pm$ 10.1} & \textbf{\underline{90.2 $\pm$ 10.1}} & 100.0 $\pm$ 0.0 \\
\hline
HHAR 1 $\rightarrow$ 3 & 77.8 $\pm$ 4.4 & 85.1 $\pm$ 3.9 & 81.3 $\pm$ 10.6 & \textbf{93.2 $\pm$ 1.6} & 90.8 $\pm$ 2.0 & 99.2 $\pm$ 0.0 \\
HHAR 3 $\rightarrow$ 5 & 68.8 $\pm$ 5.2 & 85.4 $\pm$ 1.4 & 82.3 $\pm$ 5.9 & \textbf{95.6 $\pm$ 0.9} & 94.3 $\pm$ 1.2 & 99.0 $\pm$ 0.1 \\
HHAR 4 $\rightarrow$ 5 & 60.4 $\pm$ 3.0 & 70.4 $\pm$ 3.1 & 71.6 $\pm$ 3.1 & \underline{94.2 $\pm$ 1.1} & \textbf{\underline{94.7 $\pm$ 0.5}} & 99.0 $\pm$ 0.1 \\
HHAR 0 $\rightarrow$ 6 & 33.6 $\pm$ 2.2 & 33.4 $\pm$ 1.8 & 35.6 $\pm$ 5.1 & \textbf{\underline{76.7 $\pm$ 1.5}} & \underline{74.2 $\pm$ 1.1} & 98.8 $\pm$ 0.1 \\
HHAR 1 $\rightarrow$ 6 & 72.1 $\pm$ 3.9 & 81.7 $\pm$ 3.0 & 74.9 $\pm$ 7.2 & 90.5 $\pm$ 0.7 & \textbf{90.8 $\pm$ 0.2} & 98.8 $\pm$ 0.1 \\
HHAR 4 $\rightarrow$ 6 & 48.0 $\pm$ 2.6 & 64.6 $\pm$ 5.6 & 62.7 $\pm$ 10.3 & \textbf{93.7 $\pm$ 0.4} & 85.3 $\pm$ 10.6 & 98.8 $\pm$ 0.1 \\
HHAR 5 $\rightarrow$ 6 & 65.1 $\pm$ 6.9 & 54.4 $\pm$ 1.1 & 60.0 $\pm$ 2.8 & \underline{90.7 $\pm$ 2.3} & \textbf{\underline{91.7 $\pm$ 0.4}} & 98.8 $\pm$ 0.1 \\
HHAR 2 $\rightarrow$ 7 & 49.4 $\pm$ 2.1 & 46.4 $\pm$ 3.0 & 45.0 $\pm$ 12.2 & \textbf{58.1 $\pm$ 4.5} & 56.6 $\pm$ 3.4 & 98.5 $\pm$ 0.5 \\
HHAR 3 $\rightarrow$ 8 & 77.8 $\pm$ 2.1 & 82.8 $\pm$ 1.4 & 82.2 $\pm$ 1.7 & \underline{93.4 $\pm$ 0.4} & \textbf{\underline{94.3 $\pm$ 1.0}} & 99.3 $\pm$ 0.0 \\
HHAR 5 $\rightarrow$ 8 & 95.3 $\pm$ 0.4 & 82.5 $\pm$ 2.6 & 87.5 $\pm$ 0.9 & \textbf{\underline{97.1 $\pm$ 0.3}} & \underline{95.8 $\pm$ 0.2} & 99.3 $\pm$ 0.0 \\
\hdashline
HHAR Average & 64.8 $\pm$ 16.9 & 68.7 $\pm$ 17.6 & 68.3 $\pm$ 16.4 & \textbf{\underline{88.3 $\pm$ 11.4}} & \underline{86.8 $\pm$ 11.8} & 99.0 $\pm$ 0.3 \\
\hline
WISDM AR 1 $\rightarrow$ 11 & 71.7 $\pm$ 0.0 & 55.6 $\pm$ 6.4 & 55.0 $\pm$ 11.6 & 71.7 $\pm$ 0.0 & \textbf{\underline{93.3 $\pm$ 0.0}} & 98.3 $\pm$ 0.0 \\
WISDM AR 3 $\rightarrow$ 11 & 6.7 $\pm$ 4.9 & 28.9 $\pm$ 7.5 & 45.0 $\pm$ 4.9 & \textbf{47.8 $\pm$ 0.8} & 46.7 $\pm$ 0.0 & 98.3 $\pm$ 0.0 \\
WISDM AR 4 $\rightarrow$ 15 & 78.2 $\pm$ 4.5 & 69.2 $\pm$ 5.7 & \textbf{82.7 $\pm$ 2.7} & 81.4 $\pm$ 8.9 & 75.6 $\pm$ 6.3 & 100.0 $\pm$ 0.0 \\
WISDM AR 2 $\rightarrow$ 25 & 81.1 $\pm$ 2.8 & 57.8 $\pm$ 5.5 & 72.2 $\pm$ 10.3 & 90.6 $\pm$ 1.6 & \textbf{97.8 $\pm$ 0.8} & 100.0 $\pm$ 0.0 \\
WISDM AR 25 $\rightarrow$ 29 & 47.1 $\pm$ 8.2 & 61.6 $\pm$ 5.4 & 81.9 $\pm$ 2.7 & 74.6 $\pm$ 7.4 & \textbf{84.8 $\pm$ 1.8} & 95.7 $\pm$ 0.0 \\
WISDM AR 7 $\rightarrow$ 30 & 62.5 $\pm$ 0.0 & 41.7 $\pm$ 5.1 & 61.9 $\pm$ 4.7 & \textbf{73.2 $\pm$ 16.2} & 70.2 $\pm$ 9.9 & 100.0 $\pm$ 0.0 \\
WISDM AR 21 $\rightarrow$ 31 & 57.1 $\pm$ 0.0 & 61.0 $\pm$ 8.8 & 68.6 $\pm$ 8.1 & 68.6 $\pm$ 4.0 & \textbf{\underline{92.4 $\pm$ 1.3}} & 97.1 $\pm$ 0.0 \\
WISDM AR 2 $\rightarrow$ 32 & 60.1 $\pm$ 9.1 & 49.0 $\pm$ 16.2 & 66.7 $\pm$ 4.2 & 67.3 $\pm$ 0.9 & \textbf{68.6 $\pm$ 1.6} & 100.0 $\pm$ 0.0 \\
WISDM AR 1 $\rightarrow$ 7 & 68.5 $\pm$ 2.3 & 44.8 $\pm$ 5.6 & 63.0 $\pm$ 6.0 & \textbf{70.9 $\pm$ 0.0} & 66.1 $\pm$ 6.9 & 96.4 $\pm$ 0.0 \\
WISDM AR 0 $\rightarrow$ 8 & 34.7 $\pm$ 9.3 & 13.3 $\pm$ 2.5 & 14.7 $\pm$ 8.1 & 54.0 $\pm$ 15.6 & \textbf{62.0 $\pm$ 15.7} & 99.3 $\pm$ 0.9 \\
\hdashline
WISDM AR Average & 56.8 $\pm$ 21.3 & 48.3 $\pm$ 16.0 & 61.2 $\pm$ 18.9 & 70.0 $\pm$ 11.6 & \textbf{\underline{75.8 $\pm$ 15.4}} & 98.5 $\pm$ 1.6 \\
\hline
uWave 2 $\rightarrow$ 5 & 86.3 $\pm$ 1.8 & 33.3 $\pm$ 12.0 & 18.5 $\pm$ 8.4 & 83.6 $\pm$ 12.1 & \textbf{\underline{98.2 $\pm$ 1.9}} & 100.0 $\pm$ 0.0 \\
uWave 3 $\rightarrow$ 5 & 82.7 $\pm$ 1.1 & 63.7 $\pm$ 5.3 & 32.4 $\pm$ 14.3 & \textbf{\underline{93.8 $\pm$ 5.1}} & \underline{92.9 $\pm$ 2.5} & 100.0 $\pm$ 0.0 \\
uWave 4 $\rightarrow$ 5 & 83.3 $\pm$ 0.4 & 35.4 $\pm$ 19.2 & 12.8 $\pm$ 0.4 & \textbf{\underline{99.1 $\pm$ 0.7}} & 90.2 $\pm$ 0.7 & 100.0 $\pm$ 0.0 \\
uWave 2 $\rightarrow$ 6 & 86.0 $\pm$ 0.8 & 34.5 $\pm$ 13.1 & 25.3 $\pm$ 16.2 & \textbf{\underline{93.8 $\pm$ 1.5}} & \underline{91.4 $\pm$ 1.1} & 100.0 $\pm$ 0.0 \\
uWave 1 $\rightarrow$ 7 & 95.2 $\pm$ 1.1 & 26.8 $\pm$ 13.1 & 29.2 $\pm$ 23.6 & \textbf{98.5 $\pm$ 0.4} & 91.1 $\pm$ 4.1 & 100.0 $\pm$ 0.0 \\
uWave 2 $\rightarrow$ 7 & 85.1 $\pm$ 2.2 & 53.9 $\pm$ 27.4 & 12.2 $\pm$ 0.4 & 91.4 $\pm$ 6.6 & \textbf{98.2 $\pm$ 0.7} & 100.0 $\pm$ 0.0 \\
uWave 3 $\rightarrow$ 7 & 95.5 $\pm$ 0.7 & 64.0 $\pm$ 4.9 & 30.4 $\pm$ 23.4 & 92.0 $\pm$ 8.8 & \textbf{97.6 $\pm$ 1.5} & 100.0 $\pm$ 0.0 \\
uWave 1 $\rightarrow$ 8 & \textbf{100.0 $\pm$ 0.0} & 78.6 $\pm$ 9.1 & 11.0 $\pm$ 2.9 & \textbf{100.0 $\pm$ 0.0} & 93.8 $\pm$ 3.2 & 100.0 $\pm$ 0.0 \\
uWave 4 $\rightarrow$ 8 & \textbf{100.0 $\pm$ 0.0} & 44.0 $\pm$ 25.6 & 12.5 $\pm$ 0.0 & 96.7 $\pm$ 1.1 & 93.2 $\pm$ 5.9 & 100.0 $\pm$ 0.0 \\
uWave 7 $\rightarrow$ 8 & \textbf{95.2 $\pm$ 0.4} & 49.7 $\pm$ 20.4 & 12.5 $\pm$ 0.0 & 93.8 $\pm$ 4.8 & \textbf{95.2 $\pm$ 2.3} & 100.0 $\pm$ 0.0 \\
\hdashline
uWave Average & 91.0 $\pm$ 6.5 & 48.4 $\pm$ 15.8 & 19.7 $\pm$ 8.3 & \textbf{94.3 $\pm$ 4.6} & \underline{94.2 $\pm$ 2.9} & 100.0 $\pm$ 0.0 \\
\bottomrule
\end{tabular}
\end{small}
\caption{Target classification accuracy (source $\rightarrow$ target, mean $\pm$ std\%) on 10 randomly-chosen problems for each dataset, adapting between users. The results of our proposed CoDATS model and DA-WS method are underlined when statistically significantly better (paired $t$-test with $p<0.05$) than both R-DANN and VRADA. The highest accuracy in each row is bold.}
\label{table:ssda}
\end{table*}

\begin{table}
\centering
\begin{small}
\begin{tabular}{cccc}
\toprule
Method & \shortstack{HAR 2$\rightarrow$11 \\ $H=128$} & \shortstack{HHAR 1$\rightarrow$3 \\ $H=128$} & \shortstack{uWave 2$\rightarrow$5 \\ $H=315$} \\
\midrule
No Adaptation & 0.016 $\pm$ 0.020 & 0.016 $\pm$ 0.026 & 0.036 $\pm$ 0.022 \\
R-DANN & 0.145 $\pm$ 0.029 & 0.143 $\pm$ 0.032 & 0.307 $\pm$ 0.057 \\
VRADA & 0.478 $\pm$ 0.050 & 0.452 $\pm$ 0.054 & 1.057 $\pm$ 0.048 \\
\textit{CoDATS} & \textbf{0.029 $\pm$ 0.032} & \textbf{0.029 $\pm$ 0.032} & \textbf{0.067 $\pm$ 0.031} \\
\textit{CoDATS-WS} & \textbf{0.029 $\pm$ 0.039} & \textbf{0.029 $\pm$ 0.036} & 0.068 $\pm$ 0.039 \\
\bottomrule
\end{tabular}
\end{small}
\caption{Training times per iteration (seconds) on a Nvidia Tesla K80 on three of the datasets, arranged by time series length. The lowest training time in each column is bold.}
\label{table:timing}
\end{table}

\section{Results and Discussion}

Here we present experimental results along multiple dimensions.

\subsection{Single-Source CoDATS}

Table~\ref{table:ssda} summarizes target domain classifier performance on all the benchmark datasets for single-source domain adaptation. To compare training cost, the training times are listed in Table~\ref{table:timing}.

\vspace{0.5ex}

\noindent {\bf Lower and Upper Bounds.} As expected, the lower bound (\texttt{No Adaptation}) performs poorly, providing evidence that target domain data differ from source domain data. Though, two notable exceptions are uWave 1 to 8 and uWave 4 to 8, where the lower bound reaches 100\%. In this case, adaptation should not yield a negative effect. On uWave 1 to 8, CoDATS does achieve 100\% accuracy. However, on uWave 4 to 8, it is close but not quite perfect accuracy -- though, it is far closer than either R-DANN or VRADA. This finding may indicate that CoDATS is more resistant to negative transfer than previous approaches (though further exploration is necessary). As for the upper bound, there is nearly always a sufficient amount of labeled target data to achieve close to 100\% accuracy.

\vspace{0.5ex}

\noindent {\bf Prior Methods.} R-DANN and VRADA exhibit lower performance on the uWave dataset than the others. While the other datasets contain only 128 time steps, uWave contains 315. This result may indicate that RNN models experience difficulty adapting long time series. Also, on average, R-DANN accuracy on WISDM AR and uWave is lower than that of the (approximate) lower bound. Similarly, VRADA accuracy on uWave is lower than the lower bound. These results indicate that even using the CoDATS model without applying any adaptation sometimes yields an improvement over prior work for time series sensor data.

\vspace{0.5ex}

\noindent {\bf CoDATS vs. Prior Methods.} We compare CoDATS with prior methods in terms of both accuracy and training-time efficiency. The CoDATS model improves in accuracy and consistency over the lower bound and prior methods. On average and on all but 4 of the 40 adaptation problems (one of which is a tie), CoDATS outperformed both of the prior works. On average and on all but 8 of the 40 adaptation problems (two of which are ties), CoDATS outperformed the lower bound. Since the prior baselines also use a domain-adversarial method for adaptation, the results indicate that the performance gains stem from our proposed model architecture. In addition to better performance, CoDATS typically exhibits greater consistency across random initializations as shown by the lower standard deviations on average and for all but 14 of the 40 adaptation problems than both R-DANN and VRADA.

In addition to the improvements in accuracy, Table~\ref{table:timing} indicates that CoDATS also greatly reduces training time. While the difference between CoDATS and the baseline methods depends to an extent on the number of time steps, CoDATS demonstrates faster training than R-DANN and VRADA on all datasets. We note that CoDATS requires only 20-22\% of the R-DANN training time and only 6\% of the VRADA training time for these datasets. Interestingly, CoDATS is less than twice the training time of the lower bound, indicating that adaptation can be performed with a large increase in accuracy while minimally impacting training time.

\subsection{Multi-Source CoDATS}

\begin{figure}
\centering
\subfigure[(a)][HAR]{\begin{minipage}{0.49\linewidth}
    \includegraphics[width=1.0\linewidth]{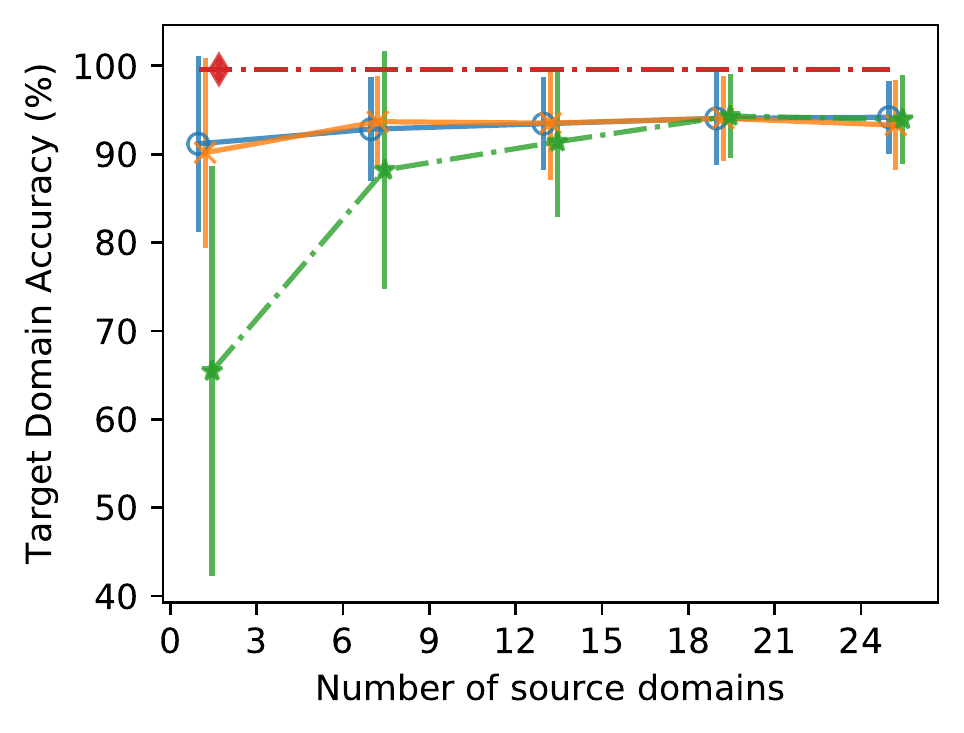}
\end{minipage}}
\subfigure[(b)][HHAR]{\begin{minipage}{0.49\linewidth}
    \includegraphics[width=1.0\linewidth]{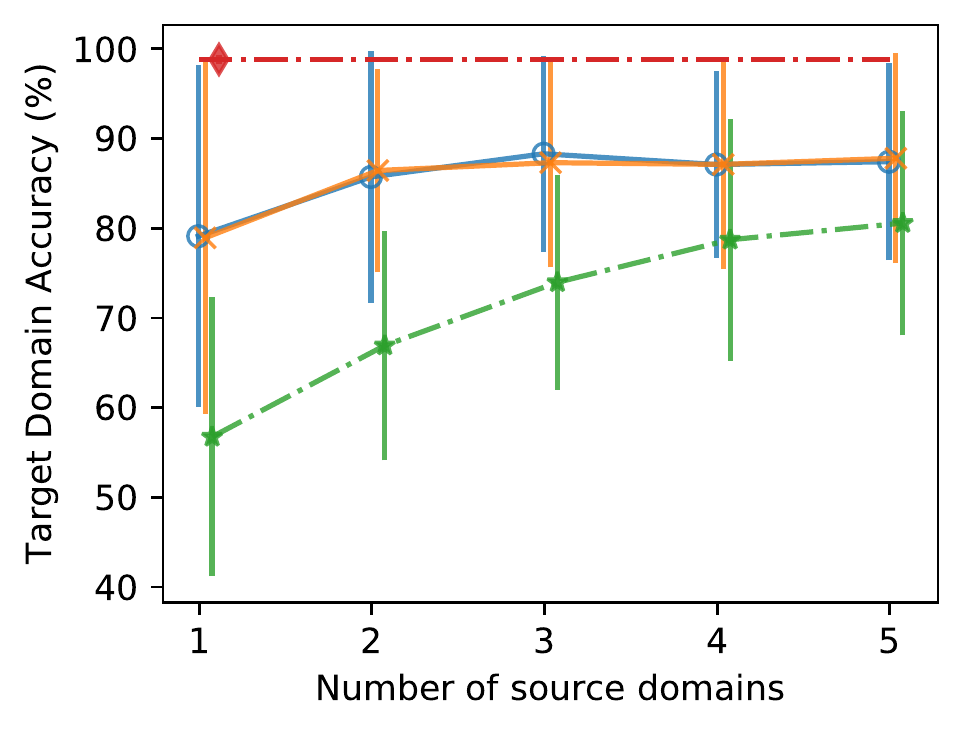}
\end{minipage}}
\subfigure[(c)][uWave]{\begin{minipage}{0.49\linewidth}
    \includegraphics[width=1.0\linewidth]{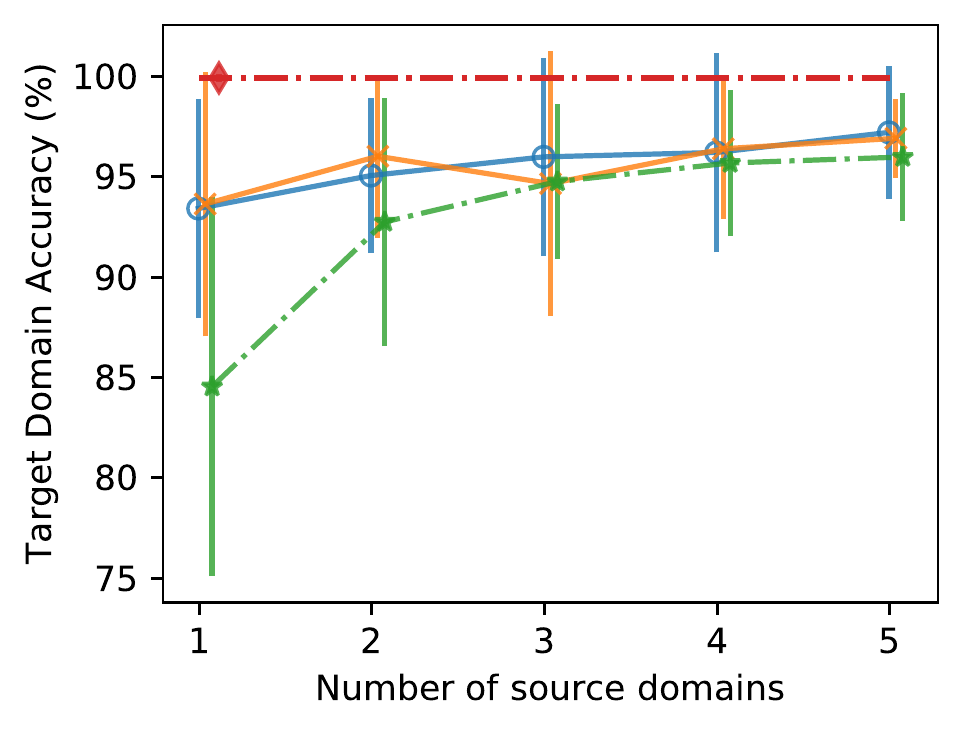}
\end{minipage}}
\subfigure[(d)][WISDM AR]{\begin{minipage}{0.49\linewidth}
    \includegraphics[width=1.0\linewidth]{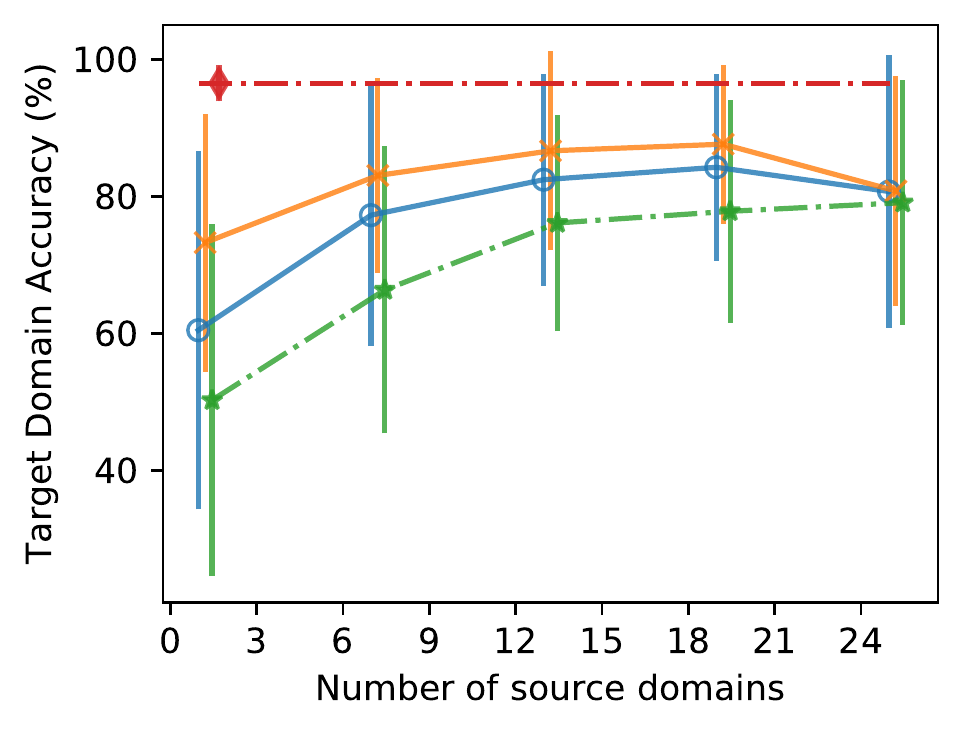}
\end{minipage}}
\includegraphics[width=1.0\linewidth]{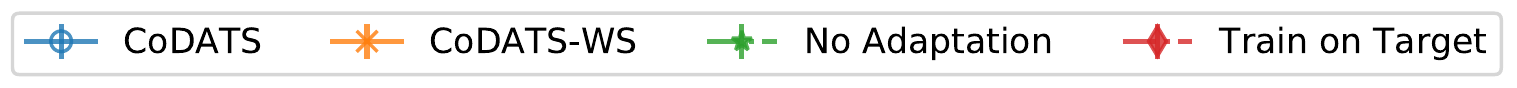}
\Description{Plots of target domain accuracy vs. number of source domains for the HAR, HHAR, uWave, and WISDM AR datasets. Four curves per plot -- CoDATS, CoDATS-WS, No Adaptation, and Train on Target.}
\caption{Target classification accuracy when varying the number of source domains $n$ (best viewed in color).}
\label{fig:vary_n_target}
\end{figure}

Figure~\ref{fig:vary_n_target} summarizes target domain classification performance as a function of $n$, the number of source domains. For each data point, we averaged over 10 random target domains for each data point in addition to 3 different random subsets of source domains.

\vspace{0.5ex}

\noindent {\bf Lower and Upper Bounds.} In agreement with the single-source results, the lower bound performs poorly with one source domain. Given additional source domains, its performance improves, indicating that additional domains better cover the space of the possible ways to perform each activity/gesture and thus improve model performance. Also as in single-source adaptation, the upper bound again yields nearly 100\% accuracy on all targets for all datasets.

\vspace{0.5ex}

\noindent{\bf CoDATS vs. Lower/Upper Bounds.} The gap between the lower bound and CoDATS is much larger at small values of $n$, as the benefit of performing domain adaptation by utilizing target domain unlabeled data has diminishing returns once more labeled source domains become available. For the highest value of $n$ on HHAR and uWave, domain adaptation still outperforms the lower bound.

The slight downturn of the CoDATS curve in HHAR and the more noticeable one in WISDM AR may stem from negative transfer. Additional domains better cover the space of possible ways of performing each activity/gesture, which yields a monotonically increasing lower bound. However, we are performing domain adaptation by learning a domain-invariant feature representation. A few people may perform activities highly different than others, meaning that if we force the representation to be invariant with respect to all of the domains, our representation may lose some ability to be discriminative. Note though that there is no downturn on the other two datasets, and the most significant drop is in the case of WISDM AR only after over 20 source domains. Thus, it is likely that in a human activity or gesture recognition settings where domain adaptation is applied, there may be fewer than 20 participants and this effect may not be noticed.

Regardless, performing multi-source adaptation on average always increases performance when compared with using only a single source -- when looking from left to right, the CoDATS curves rise, always staying above the left-most point that represents single-source adaptation performance. Thus, not only does CoDATS improve over prior single-source work, but it also facilitates utilizing data from multiple sources to further improve accuracy.

\begin{table}
\centering
\begin{small}
\begin{tabular}{cccc}
\toprule
Number of Sources & No Adaptation & \textit{CoDATS} & \textit{CoDATS-WS} \\
\midrule
$n = 1$ & 50.3 $\pm$ 25.7 & 60.5 $\pm$ 26.1 & \textbf{73.3 $\pm$ 18.8} \\
$n = 7$ & 66.4 $\pm$ 20.9 & 77.3 $\pm$ 19.1 & \textbf{83.1 $\pm$ 14.2} \\
$n = 13$ & 76.2 $\pm$ 15.8 & 82.5 $\pm$ 15.4 & \textbf{86.7 $\pm$ 14.5} \\
$n = 19$ & 77.9 $\pm$ 16.3 & 84.3 $\pm$ 13.6 & \textbf{87.7 $\pm$ 11.6} \\
$n = 25$ & 79.1 $\pm$ 17.8 & 80.7 $\pm$ 19.9 & \textbf{80.8 $\pm$ 16.8} \\
\bottomrule

\end{tabular}
\end{small}
\caption{CoDATS-WS improvement over CoDATS for classification of WISDM. The highest accuracy in each row is bold.}
\label{table:ms_daws_wisdm_ar}
\end{table}

\subsection{Domain Adaptation with Weak Supervision}

While domain adaptation with weak supervision (DA-WS) represents a different scenario than single/multi-source domain adaptation due to the availability of label proportions, we include the CoDATS-WS results in Table~\ref{table:ssda} and Figure~\ref{fig:vary_n_target} to facilitate comparing with CoDATS, which does not have label information available. A detailed view of the WISDM AR results from Figure~\ref{fig:vary_n_target} is shown in Table~\ref{table:ms_daws_wisdm_ar}.

\vspace{0.5ex}

\noindent{\bf Single-Source Domain Adaptation.}
We expect that CoDATS-WS will offer limited benefit when the source and target domains are class-balanced as for the HAR, HHAR, and uWave datasets, because CoDATS-WS capitalizes on the target label distribution differing from the source. However, it should not yield a performance degradation. Additionally, it should increase in performance on datasets with high class imbalances in the target data and between domains as in the WISDM dataset (see the Appendix for plots of class balance). The largest performance difference is on WISDM, with a 5.8\% increase. For the already-balanced datasets, CoDATS-WS yields a slight improvement on HAR (+1.8\%), a slight degradation on HHAR (-1.5\%), and nearly equal performance on uWave (-0.1\%). This indicates that in situations where target-domain label proportions are available, CoDATS-WS can improve accuracy, with the same training efficiency (Table~\ref{table:timing}).

\vspace{0.5ex}

\noindent{\bf Multi-Source Domain Adaptation.}
While using multi-source domain adaptation yields an improvement over single-source domain adaptation, CoDATS-WS combined with multi-source domain adaptation performs even better. On WISDM AR, CoDATS-WS yields a model accuracy improvement over CoDATS for all values of $n$: an increase of 12.8\% for $n$=1, 5.8\% for $n$=7, 4.2\% for $n$=13, 3.4\% for $n$=19, and 0.1\% for $n$=25, as shown in Table~\ref{table:ms_daws_wisdm_ar}. Aligning with intuition, CoDATS-WS and CoDATS perform equally on the other three balanced datasets, as shown in Figure~\ref{fig:vary_n_target}. Thus, it is evident that CoDATS-WS can yield accuracy improvements on datasets exhibiting non-uniform class distributions, which is common in real-world data.

\section{Conclusions and Future Work}
In this paper, we introduced CoDATS, a model architecture that supports time series domain adaptation. From the experimental results, it is clear that this new time series model architecture improves over prior time series adaptation work in terms of both accuracy and training time efficiency. We also demonstrated that additional accuracy gains can be achieved by utilizing data from multiple source domains and by utilizing weak supervision data from known target-domain label proportions. Future work includes extending CoDATS to support heterogeneous feature sets, developing methods to handle additional forms of weak supervision relevant for time series data, and further model improvements such as incorporating insights from InceptionTime \cite{fawaz2019inceptiontime}.

\begin{acks}
This material is based upon work supported by the \grantsponsor{nsf}{National Science Foundation}{https://www.nsf.gov/} under Grant No. \grantnum{nsf}{1543656} and by the \grantsponsor{nih}{National Institutes of Health}{https://www.nih.gov/} under Grant No. \grantnum{nih}{R01EB009675}. This research used resources from the Center for Institutional Research Computing at Washington State University.
\end{acks}

\bibliographystyle{ACM-Reference-Format}
\bibliography{bibliography}

\appendix
\newpage

\section{Reproducibility}
Here we provide additional details to aid in reproducing our results.

\subsection{Experimental Setup}
We train each model for 30,000 iterations using the Adam optimization algorithm with a learning rate of 0.0001 and a batch size of 128. We employ the DANN learning rate schedule \cite{ganin2016jmlr} for adversarial training. Because VRADA and R-DANN have no publicly available code, we use our own implementation for each method as per the details in the original papers.

We split the data from each dataset into training, validation, and test sets. The training-test split was 80\% and 20\% respectively, and the training data was further split into training-validation with the same proportions. The datasets were stratified by the labels to maintain the same label proportions for training, validation, and testing sets. Each method we evaluate only has access to the training dataset: labeled data for the sources and unlabeled data for the target. The test data is only used for the final evaluation used to create our plots. All data is normalized to have zero mean and unit variance based on statistics computed from just the training set.

Each dataset consists of data from a number of participants. In the single-source experiments, we randomly select 10 of the possible adaptation problems between two domains (excluding adapting a domain to itself). For each data point, we average and compute standard deviation over three different random initializations of the network weights on the holdout test set. In the multi-source experiments, we vary the number of source domains when adapting to a separate target domain. For each data point, we average and compute standard deviation over three different random subsets of source domains on the holdout test set. Since these are trained separately, they also are trained with three different random initializations of the network weights. Then, we average over 10 different random target domains. We select more targets than source domain subsets because we expect there to be more variance among individual domains than among random subsets of multiple domains.

During training, we perform model selection by picking the model that performs best on the holdout validation set. We evaluate the model every 4,000 iterations in addition to at the end of training. The models are trained for a total of 30,000 iterations, meaning the best model is selected from a total of 9 models. The reported accuracies reflect evaluating models that performed best on this holdout validation set. Since unsupervised domain adaptation does not use any labeled target data, these results can be interpreted as an approximate upper bound on how well these methods can perform. However, labeled target data can be employed in this manner for tuning and model selection \cite{wilson2019survey,kumar2018nips,shu2018vada}.

For single-source adaptation, DANN splits the batch size: half for the source domain and half for the target domain \cite{ganin2016jmlr}. However, for multi-source domain adaptation, we have additional source domains. We divided the batch size evenly among all domains, thereby weighting domains uniformly. On the other hand, for domain adaptation with weak supervision (DA-WS), our method depends on estimating predicted label proportions in each batch, which requires a sufficient number of target domain predictions. For the DA-WS multi-source experiments, if we divide the batch size equally among all domains, the number of target domain predictions decreases with increasing $n$, yielding an extreme decrease in performance. Thus, we instead evenly split half of the batch size among the source domains and the other half for the target domain. Note another method that would fix this issue is gradient accumulation.

\begin{figure}
    \subfigure[(a)][UCI HAR]{\begin{minipage}{1.0\linewidth}
        \centering
        \includegraphics[width=1.0\linewidth]{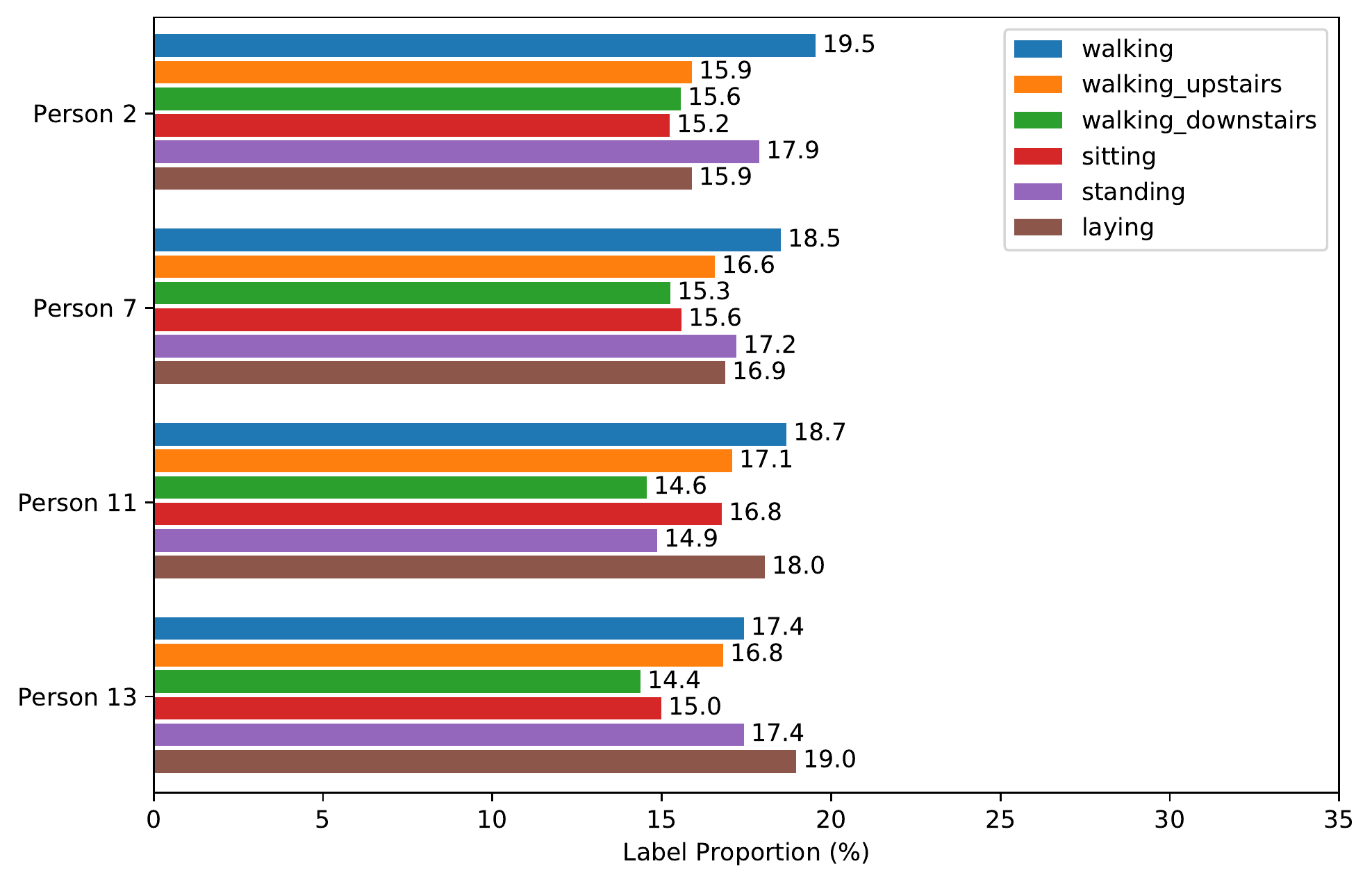}
    \end{minipage}}
    \subfigure[(b)][UCI HHAR]{\begin{minipage}{1.0\linewidth}
        \centering
        \includegraphics[width=1.0\linewidth]{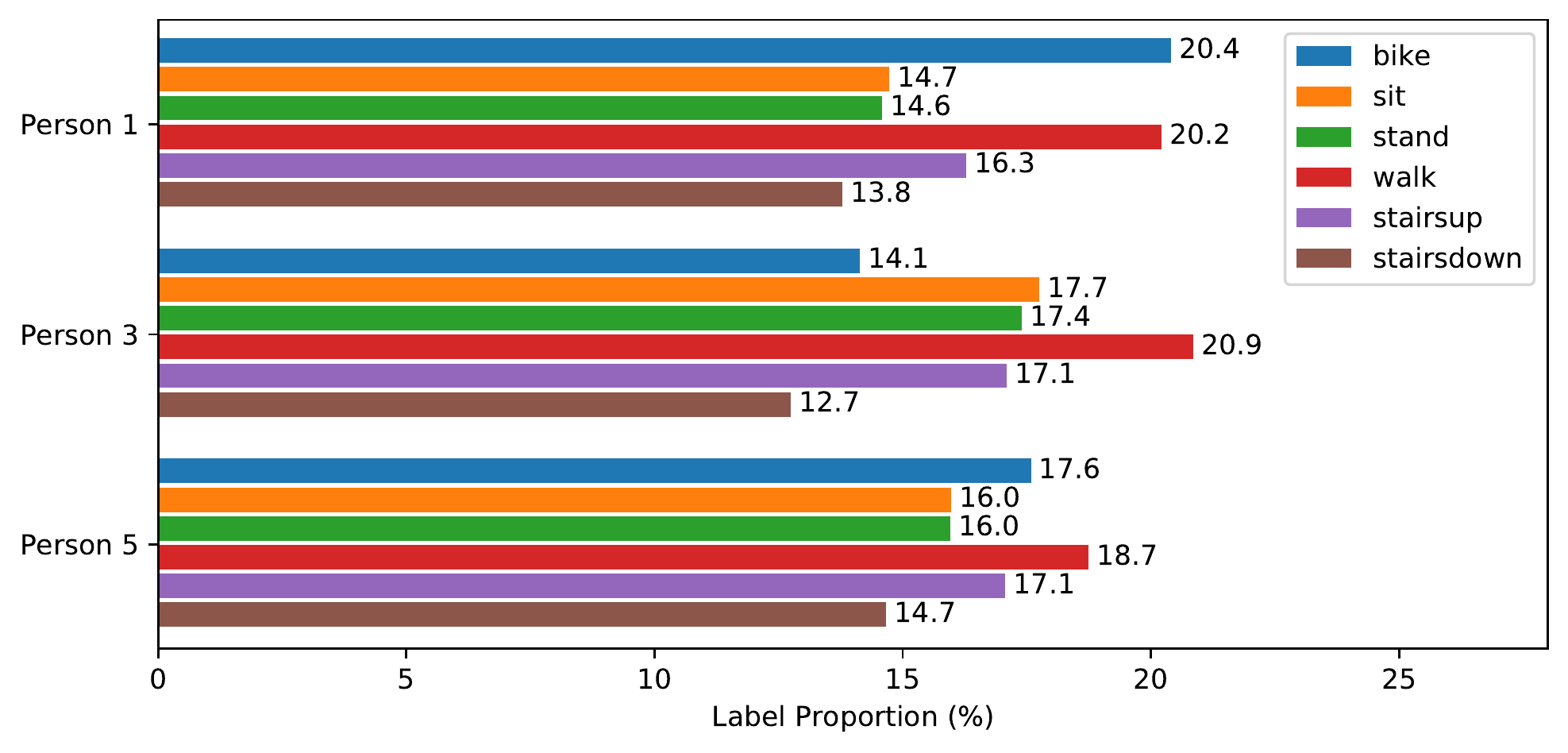}
    \end{minipage}}
    \subfigure[(c)][uWave]{\begin{minipage}{1.0\linewidth}
        \centering
        \includegraphics[width=1.0\linewidth]{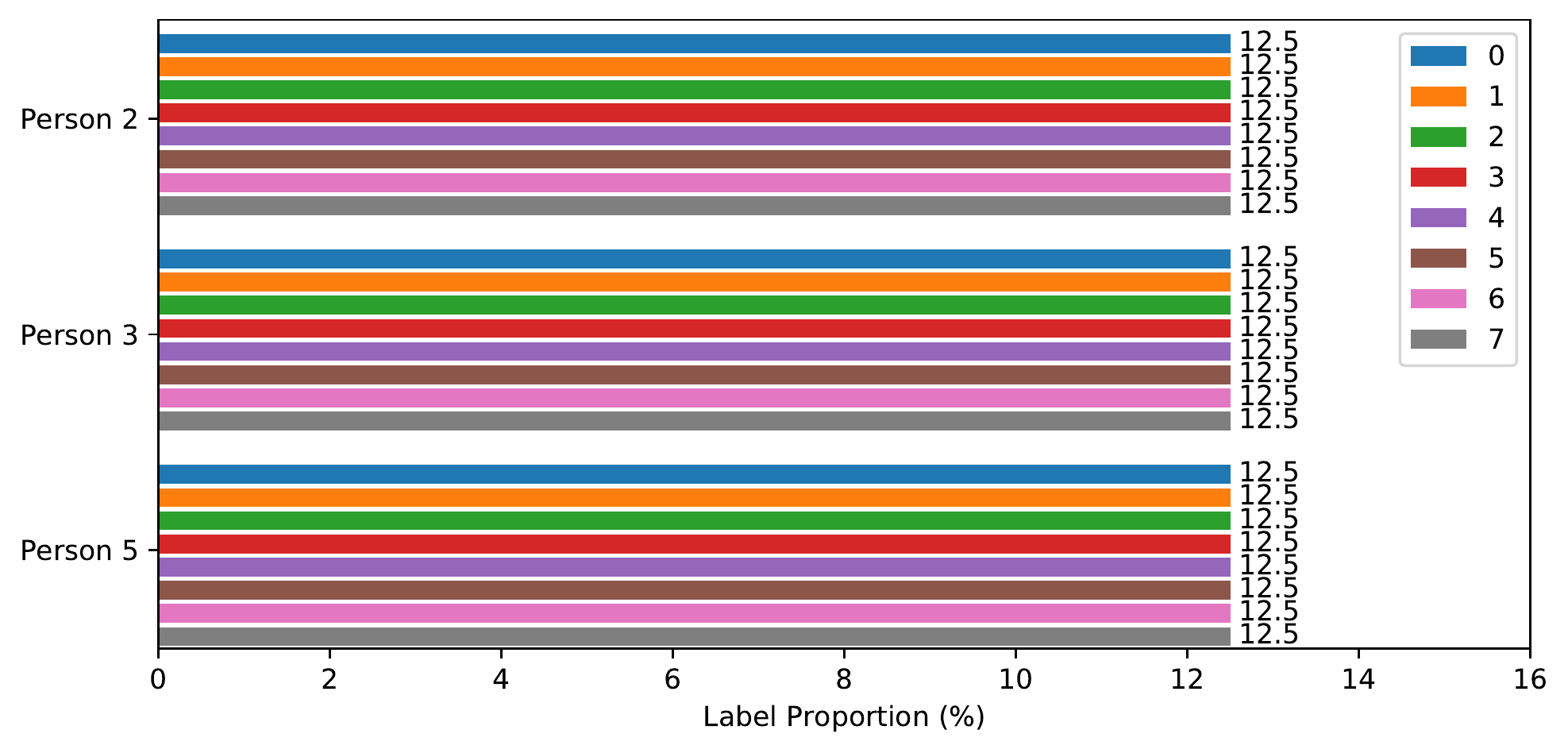}
    \end{minipage}}
    \Description{Shows each domain is similarly balanced}
    \caption{Label proportions for the participants of the relatively-balanced datasets, the first two single-source domain adaptation problems of each (best viewed in color)}
    \label{fig:class_balanced}
\end{figure}

\begin{figure}
    \centering
    \includegraphics[width=0.98\linewidth]{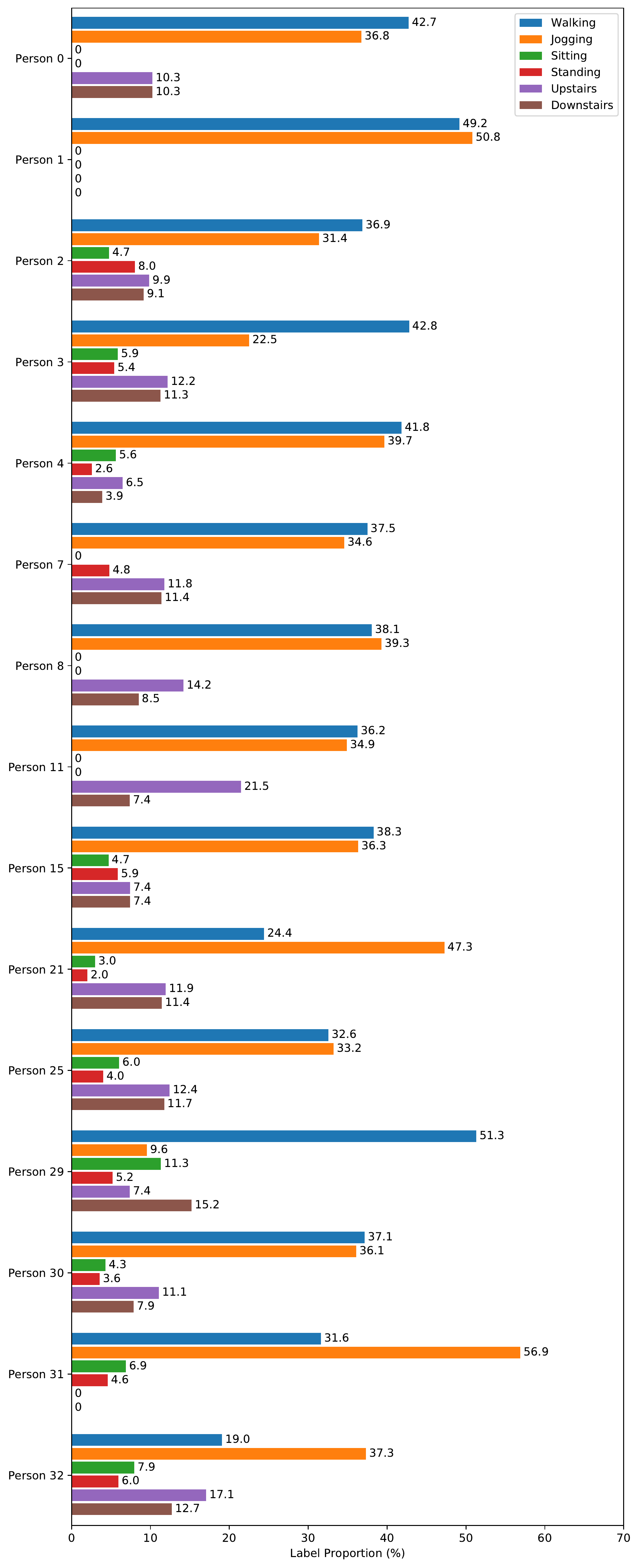}
    \Description{Shows domains have very different class balancing}
    \caption{Label proportions for the participants of the imbalanced dataset WISDM AR, domains used for single-source domain adaptation (best viewed in color)}
    \label{fig:class_unbalanced}
\end{figure}

\subsection{Datasets}
HAR \cite{anguita2013public} comes normalized and bounded to be between -1 and 1. Three-axis accelerometer, gyroscope, and estimated body acceleration were collected from 30 participants sampled at 50 Hz and come segmented into windows of 128 time steps.

The authors providing HHAR only train models with either data from one sensor or the other (not both) and found models trained with the accelerometer data to have superior performance \cite{stisen2015smartdevices}, so we similarly use the accelerometer data. The sensors were sampled at the highest rate each device would support, and we segment this data into non-overlapping windows of 128 time steps. We include the data collected from the 31 smartphones in our experiments.

For WISDM AR \cite{kwapisz2011wisdmar}, we include data from the 33 participants who have enough labeled data to yield at least 30 examples for the test set (with an 80\%-20\% train-test split). The accelerometer is sampled at 20 Hz, and we segment this into non-overlapping windows of 128 time steps.

uWave data were collected from 8 participants over 7 days, and on each day each participant performed every gesture 10 times \cite{uWaveDataset}. The data was sampled at 100 Hz. The maximum number of time steps for a gesture is 315, so we right zero-pad all gestures to this length so that all batches have a consistent size. We pad rather than segment this data because the gesture as a whole needs to be recognized, unlike in the other datasets where the activities typically consist of repetitious movements.

To show the differences in class balance, the label proportions from a few UCI HAR, UCI HHAR, and uWave participants are shown in Figure~\ref{fig:class_balanced}, which are relatively balanced between domains, and the label proportions from WISDM AR are shown in Figure~\ref{fig:class_unbalanced}, which have large differences between domains.

\end{document}